\documentclass[letterpaper, 10 pt, conference]{IEEEtran}  

\usepackage[letterpaper,top=1in,bottom=1in,left=0.75in,right=0.75in]{geometry}

\IEEEoverridecommandlockouts                              

\usepackage{booktabs}
\usepackage{multirow}
\usepackage{makecell}
\usepackage{caption}
\usepackage{array}
\usepackage{balance}
\usepackage{svg}

\usepackage{graphicx}  
\usepackage{import}    

\usepackage[font=footnotesize,labelfont=bf]{caption}
\usepackage{graphicx} 
\usepackage{hyperref}
\usepackage{booktabs}
\usepackage{multirow}
\usepackage{amsmath}
\usepackage{amssymb}
\usepackage{tabularx}
\usepackage{makecell} 
\usepackage{bbding}
\usepackage{adjustbox}  
\usepackage[style=ieee,maxnames=10,minnames=1]{biblatex}
\AtEveryBibitem{%
  \clearfield{pages}%
}
\usepackage{url} 
\hypersetup{
    colorlinks = true,
    urlcolor = blue,
    linkcolor = blue,
    citecolor = blue
} 
\title{ \textbf{DAMP: Humanoid Locomotion via Denoised Belief Learning and Adversarial Motion Priors}}

\author{Puying Shen\textsuperscript{1}, Wenhao Cui\textsuperscript{1}, Huaxing Huang, Bangyu Qin, Shengtao Li, Ziyang Dong, Guoteng Zhang%
\thanks{School of Control Science and Engineering, Shandong University.}%
\thanks{Shandong Key Laboratory of Humanoid Robotics.}%
}

\begin{document}

\maketitle
\thispagestyle{empty}
\pagestyle{empty}


\begin{abstract}
Humanoid robots possess the structural capability to traverse complex terrains. However, achieving stable t raversal without relying on perceived information remains challenging, particularly in complex environments. This paper introduces DAMP, a reinforcement learning framework aimed at achieving robust and naturalistic humanoid locomotion over challenging terrains, with the assumption that no perceived information is available. The framework leverages recurrent neural networks to capture temporal dependencies and implicitly infer privileged and other task-relevant latent information. By aligning the learned representations with the task objective, the method enables robust and goal-consistent policy learning. This end-to-end framework achieves transfer learning from simulation to real-world environments, demonstrating the proposed method's robustness and generalization capabilities. The video of the real-world demonstration can be found at the following link: https://youtu.be/AkI7TZB2DDM.
\end{abstract}


\section{INTRODUCTION}
Humanoid robots face significant challenges when traversing complex terrains such as stairs, uneven surfaces, and obstacles. Compared with quadrupeds, humanoids exhibit a higher center of mass, greater degrees of freedom, and more complex kinematic structures, which collectively compromise stability and robustness in challenging environments.

Traditional model-based control strategies, including zero-moment point control (ZMP) \cite{kajita2003biped}, model predictive control (MPC) \cite{wieber2006trajectory}, and whole-body control (WBC) \cite{sentis2006whole}, have demonstrated effective gait generation for humanoid locomotion. However, these approaches rely heavily on accurate modeling of environmental dynamics, limiting their applicability in scenarios involving complex terrain interactions such as obstacle traversal \cite{ramuzat2022benchmarking}.

Reinforcement learning (RL) provides a model-agnostic alternative that improves adaptability to unknown or dynamically changing environments. Representative frameworks, such as the Hybrid Internal Model (HIM) \cite{long2024hybrid} and DreamWaQ \cite{Made2023DreamWaQLR}, have shown strong performance in quadrupedal locomotion and can be transferred to humanoids. Estimator-based RL methods \cite{long2024hybrid,Made2023DreamWaQLR,kumar2021rma} have successfully reduced the sim-to-real gap and enabled traversal over diverse terrains. These pipelines were later extended to humanoids \cite{huang2025think,wang2024toward,cui2024adapting}, providing basic terrain-crossing capabilities. Nevertheless, purely task-driven RL often struggles to generate natural and energy-efficient motions, particularly under sparse contact events or complex terrain transitions, and still fails on high-difficulty terrains such as 10 cm continuous steps or 20 cm railings.

To address these limitations, subsequent work incorporated exteroceptive sensing, such as depth cameras and LiDAR. Methods like Perceptive Internal Model \cite{long2025learning} and Humanoid Parkour \cite{zhuang2024humanoid} enable traversal of sparse high obstacles. However, these approaches introduce additional computational overhead and are susceptible to sensor drift, limiting long-term autonomous deployment.

In parallel, imitation learning (IL) has been explored to improve motion naturalness. Methods such as adversarial motion prior (AMP) can produce smooth and human-like behaviors, but they depend on high-quality demonstration data and often exhibit limited generalization. Huang et al. \cite{huang2025think} combined HIM with AMP, achieving improved stylistic locomotion performance.

Humanoid imitation learning has further evolved through motion imitation with style priors, generative modeling, teleoperation, and multimodal feedback. Early work such as DeepMimic \cite{peng2018deepmimic} and its extensions \cite{won2020scalable} relied on dense motion tracking rewards to reproduce high-fidelity skills. AMP \cite{peng2021amp} introduced style discriminators to jointly optimize motion realism and task objectives. More recently, generative approaches such as BeyondMimic \cite{liao2025beyondmimic} leveraged diffusion models to translate human demonstrations into diverse robotic behaviors. Wen et al. \cite{wen2025constrained} further integrated elevation maps into AMP to enable perception-based locomotion in simulation.

The Denoising World Model \cite{Gu-RSS-24}, proposed by Gu et al., combines a recurrent neural network (RNN) with a denoising mechanism for observation data, enhancing the robustness of the model by mitigating noise in the input information. However, strategies relying on gait rewards still face certain difficulties in continuous obstacle crossing.

Inspired by these findings, we propose Denoised Belief Learning with Adversarial Motion Priors (DAMP), a framework for stable and task-compliant humanoid locomotion with enhanced motion naturalness. DAMP adopts a LSTM-based encoder–decoder architecture to implicitly infer latent and privileged information. During training, adversarial imitation learning and dynamic reward interpolation are jointly employed to enhance locomotion performance on challenging terrains. The entire pipeline is trained in a single-stage, end-to-end manner, resulting in a hardware-deployable policy. We deploy DAMP on the Noetix N2 \cite{noetixrobotics2025products} humanoid robot and validate its effectiveness through extensive real-world experiments over complex terrains, demonstrating smooth, robust, and natural locomotion.

\section{PRELIMINARY}\label{subsec:amp}
We formulate humanoid locomotion as a Markov Decision Process (MDP) defined by the tuple 
\(\langle \mathcal{S}, \mathcal{A}, \mathcal{T}, \mathcal{R}, \gamma \rangle\), 
where:
\begin{itemize}
    \item \(\mathcal{S}\) and \(\mathcal{A}\) denote the state and action spaces.
    \item \(\mathcal{T}(\mathbf{s}_{t+1} \mid \mathbf{s}_t, \mathbf{a}_t)\) defines the transition dynamics.
    \item \(\mathcal{R}(\mathbf{s}_t, \mathbf{a}_t)\) denotes the reward function.
    \item \(\gamma \in [0,1]\) is the discount factor.
\end{itemize}

The action $\mathbf{a}_t \in \mathcal{A}$ is sampled from the policy 
$\pi(\cdot \mid \mathbf{o}_{1:t})$. 
The policy employs an RNN to encode the observation and outputs a conditional probability distribution over the action space. 
Therefore, the locomotion problem is naturally formulated as a discrete-time partially observable Markov decision process (POMDP), where decision-making relies on partial and temporally correlated observations. 
The objective is to optimize the policy $\pi$ to maximize the expected cumulative discounted return:
\begin{equation}
J(\pi) = \mathbb{E}_{\pi}\left[ \sum_{t=0}^{\infty} \gamma^t \mathcal{R}(\mathbf{s}_t, \mathbf{a}_t) \right].
\end{equation}
Here, $\mathcal{R}(\mathbf{s}_t, \mathbf{a}_t)$ denotes the immediate reward provided by the environment.

To encourage natural and human-like locomotion, we incorporate the Adversarial Motion Prior (AMP) framework. 
Unlike methods that directly track reference trajectories, AMP guides the policy toward generating motions consistent with expert demonstrations through adversarial learning.

For adversarial training, we define a discriminator-specific state representation 
$\mathbf{s}_t^{I}$ derived from the proprioceptive observation $\mathbf{o}_t$. 
The discriminator operates on consecutive state transitions 
$(\mathbf{s}_t^{I}, \mathbf{s}_{t+1}^{I})$.

Specifically, we introduce a discriminator 
$D_\psi(\mathbf{s}_t^{I}, \mathbf{s}_{t+1}^{I})$ 
to distinguish transitions from expert demonstrations 
and those generated by the policy.

The adversarial loss is defined as:
\begin{equation}
\begin{aligned}
\mathcal{L}_{AMP}
=&\; \frac{1}{2}
\mathbb{E}_{(\mathbf{s}_t^{I},\mathbf{s}_{t+1}^{I}) \sim \mathcal{D}_E}
\left[
\left(D_\psi(\mathbf{s}_t^{I},\mathbf{s}_{t+1}^{I}) - \frac{1}{2}\right)^2
\right] \\
&+
\frac{1}{2}
\mathbb{E}_{(\mathbf{s}_t^{I},\mathbf{s}_{t+1}^{I}) \sim \mathcal{D}_\pi}
\left[
\left(D_\psi(\mathbf{s}_t^{I},\mathbf{s}_{t+1}^{I}) + \frac{1}{2}\right)^2
\right] \\
&+
\lambda_{GP}
\mathbb{E}_{\hat{\mathbf{s}} \sim \mathcal{D}_{\text{interp}}}
\left[
\|\nabla_{\hat{\mathbf{s}}} D_\psi(\hat{\mathbf{s}})\|_2^2
\right].
\end{aligned}
\end{equation}

Here, transitions are sampled from the expert dataset $\mathcal{D}_E$ 
and policy rollouts $\mathcal{D}_\pi$, respectively. 
The gradient penalty weighted by $\lambda_{GP}$ stabilizes adversarial training.

The style reward derived from the discriminator is defined as:
\begin{equation}
r_{\text{style}}(\mathbf{s}_t^{I},\mathbf{s}_{t+1}^{I})
=
\max \left(0,\, 1 - \frac{1}{4} D_\psi(\mathbf{s}_t^{I},\mathbf{s}_{t+1}^{I}) \right).
\end{equation}

The final reward combines the task objective and the adversarial motion prior:
\begin{equation}
r_t
=
\alpha\, r_{\text{task}}(\mathbf{s}_t,\mathbf{a}_t)
+
\beta\, r_{\text{style}}(\mathbf{s}_t^{I},\mathbf{s}_{t+1}^{I}).
\end{equation}

The adversarial formulation promotes motion diversity and training stability, while the dynamic reward interpolation mechanism balances task-oriented objectives with motion style consistency, improving robustness in complex environments.

\section{METHOD}

This paper proposes a single-stage end-to-end learning framework that directly generates desired joint position commands from onboard proprioceptive observations using a recurrent neural network (RNN). The framework is lightweight and capable of estimating useful information such as foothold states and velocity tracking conditions. We first provide an overview of the proposed framework and then describe it in detail.

\begin{figure*}[htbp]
    \centering
    \includegraphics[width=\linewidth]{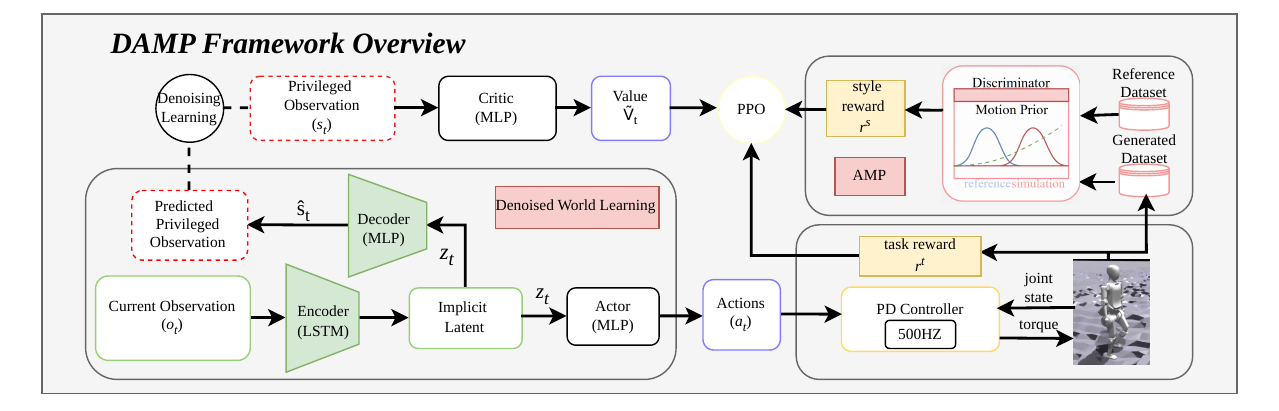}
    \caption{Overview of the DAMP framework. The LSTM receives single-frame observations and encodes them into the latent variable $z_t$. The estimation module decodes $z_t$ with privileged information $s_t$ for denoising, providing a more reliable estimate. AMP is trained jointly with PPO.}
    \label{fig-overview}
\end{figure*}

\subsection{Overview}
As illustrated in Fig.~\ref{fig-overview}, the proposed DAMP framework is built upon three principal components: denoised world learning (DWL) module, the critic network, and AMP which has been introduced in Section~\ref{subsec:amp}. During training, the DWL module processes proprioceptive observations available at deployment, while the decoder reconstructs the latent variable toward privileged information, encouraging latent alignment with richer supervisory signals. Detailed inference procedures are provided in Section~\ref{subsec:dwl}.
The critic network receives additional privileged information to provide more informative value estimation, thereby stabilizing policy optimization. The overall framework is trained using Proximal Policy Optimization (PPO), enabling end-to-end optimization of all components within a unified reinforcement learning paradigm.

\noindent\textit{1) Policy Network:}
The policy network consists of a recurrent backbone and a feedforward actor head. 
Specifically, an LSTM-based recurrent neural network encodes temporal dependencies from sequential observations, and a single-head multilayer perceptron (MLP) serves as the actor to output joint position commands.

At each timestep $t$, we define a proprioceptive observation vector
\begin{equation}
\mathbf{o}_t = \{ \mathbf{c}_t, \; \boldsymbol{\omega}_t, \; \mathbf{g}_t, \; \boldsymbol{\theta}_t, \; \dot{\boldsymbol{\theta}}_t, \; \mathbf{a}_{t-1} \},
\end{equation}
where $\mathbf{c}_t = \left( v^{x}_{t},\; v^{y}_{t},\; \omega^{\mathrm{yaw}}_{t} \right)$ denotes the commanded linear and yaw velocities, 
$\boldsymbol{\omega}_t$ is the base angular velocity measured by the IMU, 
$\mathbf{g}_t$ represents the gravity direction expressed in the body frame, 
$\boldsymbol{\theta}_t$ and $\dot{\boldsymbol{\theta}}_t$ denote joint positions and velocities obtained from joint encoders, and 
$\mathbf{a}_{t-1}$ is the action applied at the previous timestep. 
The policy takes $\mathbf{o}_t$ as input at each timestep.

\noindent\textit{2) Value Network:} The value network takes a privileged state $\mathbf{s}_t$ as input.
Unlike the policy, which only receives proprioceptive observation
$\mathbf{o}_t$, the critic has access to additional privileged
information available during training.

Specifically, the full critic state is defined as
\begin{equation}
\mathbf{s}_t =
\left\{
\mathbf{o}_t,\;
\mathbf{v}_t,\;
m_t,\;
\mu_t,\;
\rho_t,\;
\mathbf{k}_{p},\;
\mathbf{k}_{d},\;
\boldsymbol{\eta}_t,\;
\mathbf{c}^{\text{foot}}_t,\;
\mathbf{h}_t
\right\},
\end{equation}
where $\mathbf{v}_t \in \mathbb{R}^3$ denotes the base linear velocity,
$m_t$ is the payload mass,
$\mu_t$ and $\rho_t$ represent the friction and restitution coefficients,
$\mathbf{k}_{p}$ and $\mathbf{k}_{d}$ denote joint stiffness and damping factors,
$\boldsymbol{\eta}_t$ represents motor strength scaling factors,
$\mathbf{c}^{\text{foot}}_t \in \mathbb{R}^2$ indicates binary foot contact states. $\mathbf{h}_t$ denotes a terrain height scan capturing vertical elevation measurements within a $1.2\,\mathrm{m} \times 0.8\,\mathrm{m}$ rectangular area centered at the robot base.
\noindent\textit{3) Action Space:} The action $\mathbf{a}_t$ corresponds to the target positions for the robot joints.  
The joint torques $\boldsymbol{\tau}_t$ are computed using a PD controller:

\begin{equation}
\boldsymbol{\tau}_t = \mathbf{k}_p (\mathbf{a}_t - \boldsymbol{\theta}_t) - \mathbf{k}_d \, \dot{\boldsymbol{\theta}}_t,
\end{equation}

\noindent\textit{3) Reward Function:} 
We design the reward function to jointly encourage accurate command tracking, locomotion stability, energy efficiency, and physically consistent contact behaviors. Detailed reward terms and weights are summarized in \autoref{table-reward}. 

Here, $\phi_{xy}$ denotes the robot's roll and pitch angles. 
$\mathbf{v}_{\text{feet}}$ represents the foot linear velocity, 
$F_{\text{contact}}$ denotes the foot contact force with an upper bound $F_{\max}$. 
$\boldsymbol{\theta}$ and $\boldsymbol{\theta}_{\text{default}}$ denote joint positions and their default values 
($\boldsymbol{\theta}_{\text{up}}$ and $\boldsymbol{\theta}_{\text{default,up}}$ for upper-body joints). 
$\ddot{\boldsymbol{\theta}}$ represents joint accelerations, 
$\boldsymbol{\tau}$ denotes joint torques, 
$\mathbf{a}_t$ is the action at time $t$, 
$F_{\text{collision}}$ is the collision force, 
and $\mathbf{1}(\cdot)$ denotes the indicator function.

\begin{table}[t]
\centering
\caption{Humanoid Locomotion Reward Terms}\label{table-reward}
\renewcommand{\arraystretch}{1.0}
\scriptsize
\begin{tabularx}{\linewidth}{l X c}
\toprule[1.25pt]
Reward & Equation & Weight \\
\midrule
Linear velocity tracking & $\exp\big(-5\|\mathbf{v}_{xy}^\text{cmd} - \mathbf{v}_{xy}\|_2^2\big)$ & 1.2 \\
Angular velocity tracking & $\exp\big(-5(\omega_\text{yaw}^\text{cmd} - \omega_\text{yaw})^2\big)$ & 1.0 \\
Orientation stability & $\exp\Big(-5 \, \|\boldsymbol{\phi}_{xy} \|_2^2 \Big)$ & 0.5 \\
Contact with zero velocity & $\sum \| \mathbf{v}_\text{feet} \cdot \mathbf{c}^{\text{foot}} \|^2$ & -0.2 \\
Feet contact forces & $\sum (\| {F}_\text{contact} \| - F_\text{max})$ & -0.01 \\
Feet air time & $\min(t_\text{air}/t_\text{contact}, 0.5)$ & 3.0 \\
Default joint pos. & $\exp(-\|\boldsymbol{\theta} - \boldsymbol{\theta}_\text{default}\|)$ & 0.5 \\
Default upper joint pos. & $\exp(-\|\boldsymbol{\theta}_\text{up} - \boldsymbol{\theta}_\text{default,up}\|)$ & 0.5 \\
DOF accelerations & $\sum \|\ddot{\boldsymbol{\theta}}\|^2$ & -2.5e-7 \\
Energy cost & $\|\boldsymbol{\tau} \cdot \dot{\boldsymbol{\theta}}\|$ & -1e-3 \\
Action smoothness & $\sum \|\mathbf{a}_t - 2\mathbf{a}_{t-1} + \mathbf{a}_{t-2}\|^2$ & -0.01 \\
Collision & $\sum \mathbf{1}(\|\mathbf{F}_\text{collision}\|>5)$ & -50.0 \\
DOF position limits & penalize $\text{out-of-limits}$ & -5.0 \\
DOF velocity limits & penalize $\text{out-of-limits}$ & -10.0 \\
\bottomrule[1.25pt]
\end{tabularx}
\end{table}

In this work, we adopt the feet-contact function in (\ref{feetcontact}) to enhance the naturalness of robot gait, 
improving locomotion stability and smoothness compared with approaches based on no-fly penalties or joint referenced rewards based on cycle time.

\begin{equation}
\label{feetcontact}
\begin{aligned}
r_\text{feet\_contact} = \mathbf{1}\Big[ \; 
& (c_L \oplus c_R)  \lor \; (T_\text{both} < 0.2) 
 \lor \; c_\text{standing} \; \Big]
\end{aligned}
\end{equation}
where $c_L, c_R \in \{0,1\}$ denote the contact states of the left and right feet,  
$\oplus$ is the XOR operation, $T_\text{both}(t)$ is the accumulated time with both feet in contact, 
$c_\text{standing}$ indicates whether a standing command is active.

\subsection{Denoising Model}\label{subsec:dwl}

We adopt an encoder-decoder architecture to estimate and process observation data.
During training, noise is injected into the environment to simulate diverse conditions,
enabling the latent variables to be inferred more robustly while capturing such variations.
The recurrent encoder extracts the latent state $\mathbf{z}_t$, which is then decoded into a vector of the same dimension as the underlying state $\mathbf{s}_t$.
Formally, the model can be expressed as:
\begin{equation}
P(\tilde{\mathbf{s}}_t) = \mathbb{E}_{\mathbf{o}_{\le t}} \int_{\mathbf{z}}
P_{\text{Decoder}}(\tilde{\mathbf{s}}_t \mid \mathbf{z}_t)\,
P_{\text{Encoder}}(\mathbf{z}_t \mid \mathbf{o}_{1:t}) \, d\mathbf{z} ,
\label{eq:denoise_model}
\end{equation}

Since the observation space is substantially higher-dimensional than the latent space, the architecture naturally forms an information bottleneck that filters out observation noise and task-irrelevant variations. To further regularize the belief representation and enhance robustness, we impose an additional $L_1$ penalty on the latent state $\mathbf{z}_t$ to encourage sparsity. As the goal is state denoising rather than generative synthesis, a simple deterministic reconstruction loss is adopted. The resulting denoising loss is formulated as:

\begin{equation}
\mathcal{L}_{\text{denoise}} 
= \| \tilde{\mathbf{s}}_t - \mathbf{s}_t \|_2^2 + \lambda_r \| \mathbf{z}_t \|_1 ,
\label{eq:denoise_loss}
\end{equation}
Here, $\lambda_r$ regulates the strength of latent regularization. Incorporating privileged signals, including ground friction, actuator torques, and terrain elevation, facilitates accurate online adaptation and implicit system identification. During deployment, only the encoder and actor are retained, while the decoder and privileged supervision are discarded.

\subsection{Lipschitz Continuity Penalty (LCP)}

For legged robots, smooth policy output and stable policy updates are essential. To address these limitations, we employ a simple and fully differentiable smoothness objective function based on Lipschitz continuity \cite{chen2024lcp} of the policy. A function is Lipschitz continuous if its gradients are bounded, ensuring that small input perturbations produce bounded action variations. By constraining the gradient of the policy with respect to the state, we can directly enforce smooth control output during training. Formally, for a stochastic policy $\pi(\mathbf{a}|\mathbf{s})$, the gradient penalty is
\begin{equation}
\mathcal{L}_\text{GP} = \mathbb{E}_{\mathbf{s},\mathbf{a} \sim D} \big[ \|\nabla_{\mathbf{s}} \log \pi(\mathbf{a}|\mathbf{s})\|_2^2 \big],
\end{equation}
where $D$ denotes a dataset of state-action pairs $(\mathbf{s}_t, \mathbf{a}_t)$ collected from policy rollouts. This penalty encourages the policy to respect a Lipschitz constraint, reducing abrupt changes in actions across nearby states. We incorporate this loss into the PPO during training to impose certain constraints on the policy.

\subsection{Training Details}
\noindent\textit{1) Simulation Platform:} We used the NVIDIA IsaacGym simulator \cite{makoviychuk2021isaac} to train the actor, critic, and denoising models. We were able to complete 7000 iterations of training in 5 hours on an NVIDIA RTX 4090 graphics card, resulting in a network that could be deployed immediately.

\noindent\textit{2) Terrain Curriculum:} We employed a training program that progressively increased terrain difficulty to allow the strategy to gradually adapt to complex terrain. 
Our training terrain included steps with heights ranging from $8$ to $20\,\mathrm{cm}$, irregular cobblestone paths, continuous slopes of up to $15^\circ$, $20\,\mathrm{cm}$ curbs, and irregular cobblestone obstacles. The specific mechanism for determining terrain levels and assigning lessons based on level and policy performance is detailed in Subsection~\ref{subsec:terrain_adoption_rate}.
The sampling range for forward speed commands was $[0.0, 0.8]\,\mathrm{m/s}$, and the sampling range for horizontal angular velocity was $[-1, 1]\,\mathrm{rad/s}$.

\noindent\textit{3) Domain Randomization:} To enhance the robustness of policies trained in simulation and facilitate smooth sim-to-real transfer, we apply domain randomization to the robot’s simulation parameters (joint P/D-gains, motor strength, body mass, center of mass (CoM), payload, and initial joint positions) and environment properties (ground friction, restitution, external pushes, and disturbance timing). The ranges for these randomized parameters are summarized in \autoref{table-domain-rand}.

\begin{table}[t]
\centering
\caption{Domain Randomization Parameters}\label{table-domain-rand}
\renewcommand{\arraystretch}{1.1}
\begin{tabular}{lll}
\toprule[1.25pt]
Parameter & Range & Unit \\
\midrule
Joint P-gain factor & $[0.8, 1.2]$ & $N\cdot m / \mathrm{rad}$ \\
Joint D-gain factor & $[0.8, 1.2]$ & $N\cdot m / \mathrm{rad}$ \\
Motor strength & $[0.8, 1.2]$ & $N\cdot m$ \\
COM displacement & $[-0.05, 0.05]$ & m \\
Friction coefficient & $[0.1, 2.0]$ & - \\
Restitution coefficient & $[0.0, 1.0]$ & - \\
Base mass & $[-5.0, 5.0]$ & kg \\
Push force & $[50.0, 300.0]$ & N \\
Push torque & $[25.0, 100.0]$ & N·m \\
Disturbance probability & 0.002 & - \\
Disturbance interval & $[10, 25]$ & steps \\
\bottomrule[1.25pt]
\end{tabular}
\end{table}

\section{EXPERIMENTS}\label{sec:experiments}

To validate the effectiveness of our framework, we conducted ablation experiments using the Noetix N2 robot in both simulation and real-world environments. This section describes the experimental setup, followed by the presentation and analysis of the results.

\subsection{Experiment Configurations}
The N2 robot has 18 degrees of freedom, with ankle pitch as the only degree of freedom at the ankles. It stands 1.2 m tall and weighs 33 kg, and the policy runs at a control frequency of 50 Hz. 

We designed two sets of ablation experiments to compare with our own experiments:
\begin{itemize}
    \item \textit{DAMP w/o} $\mathcal{L}_{\text{denoise}}$ : This framework removes the denoising loss during policy updates.
    \item \textit{DAMP w/o} $\mathcal{L}_{AMP}$ and $r_{\text{style}}$: This framework removes the style reward and adversarial loss from imitation learning, thus eliminating the imitation paradigm.
    \item \textit{HIM}: The framework proposed in \cite{huang2025think} is included in our comparative experiments as a baseline for evaluation.
\end{itemize}

\begin{figure*}[htbp]
    \centering
    \includegraphics[width=0.75\linewidth]{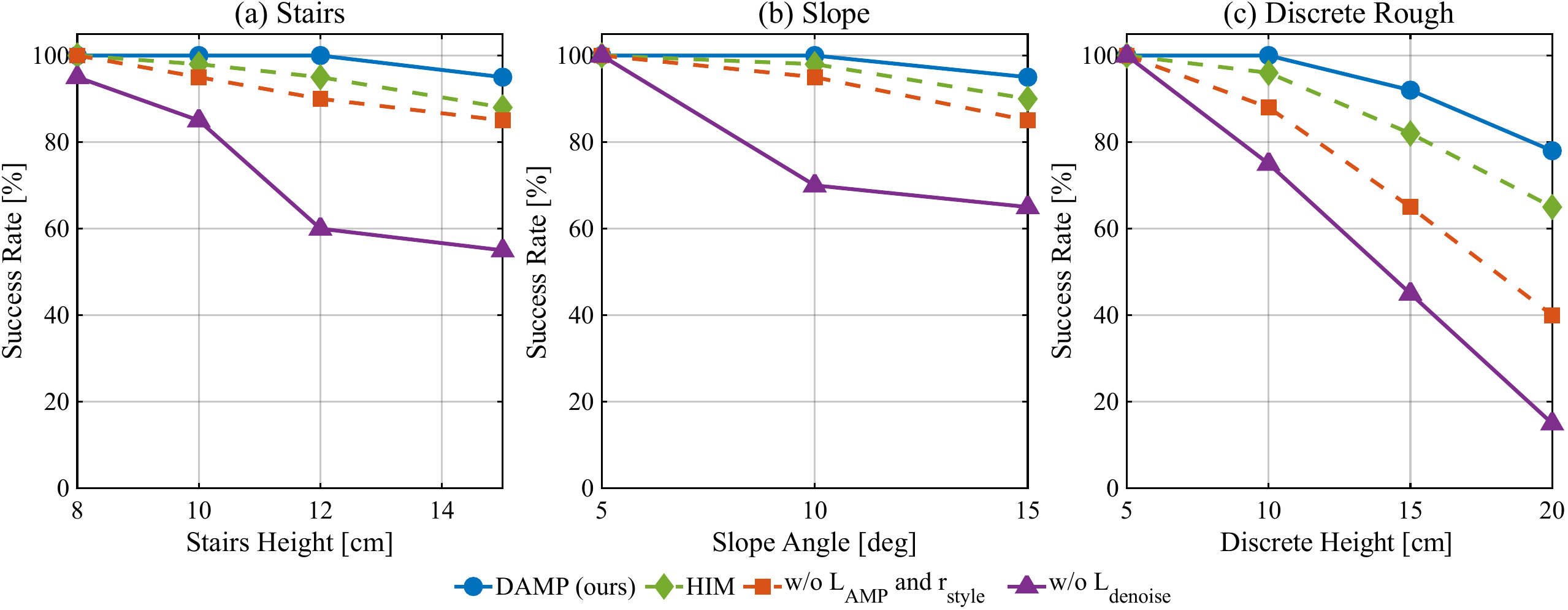}
    \caption{Traversal success rate under increasing terrain difficulty, including stairs, slopes, and discrete obstacles. DAMP (ours) consistently outperforms ablated variants, demonstrating superior robustness. Removing $\mathcal{L}_{AMP}$ and $r_{\text{style}}$ degrades stability, while removing $\mathcal{L}_{\text{denoise}}$ causes the largest performance drop.}
    \label{success_rate}
    
\end{figure*} 
\begin{figure}[htbp]
    \centering
    \includegraphics[width=\linewidth]{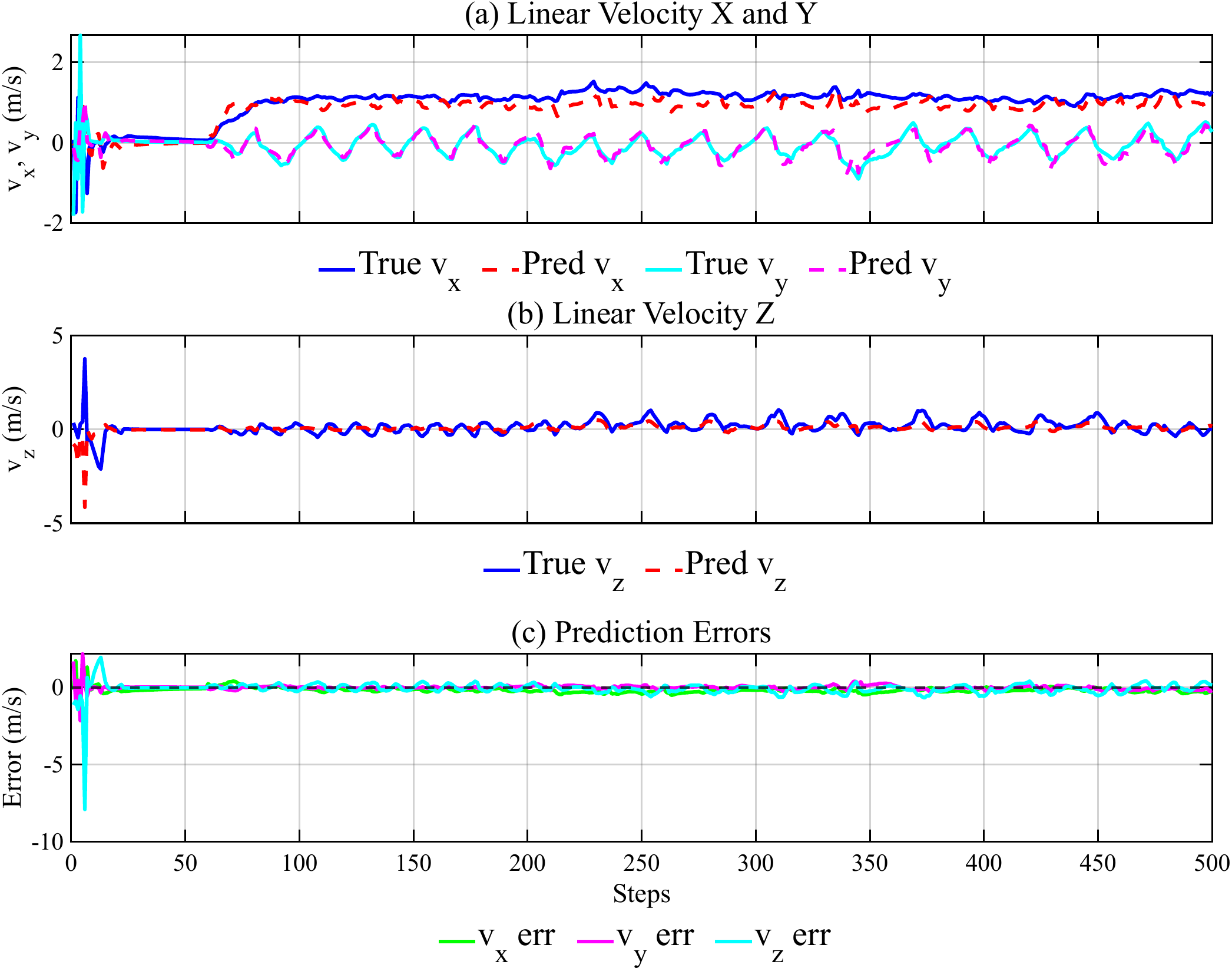}
    \caption{Comparison between the predicted linear velocity from the denoising module and the ground-truth linear velocity, along with the corresponding prediction error.}
    \label{predict_lin}
\end{figure}
\subsection{Simulation Experiments}\label{subsec:terrain_adoption_rate}
We used the terrain curriculum adoption rate as the evaluation metric to assess the robustness of the framework under different ablation configurations.
Similarly to the curriculum learning strategy used in \cite{gu2024humanoid}, the terrain difficulty in our training procedure is adjusted according to robot performance. If the robot traverses more than 50\% of the episode distance, the terrain difficulty level is increased. 
Conversely, if the robot travels less than half the required distance, the terrain becomes simpler. This dynamic adaptation allows the robot to progress through increasingly challenging terrains as its performance improves. Robots that reach the highest terrain level are reassigned to a random level, ensuring variability.
During the ablation experiments, we repeated each configuration multiple times and computed the average terrain level achieved by each framework. 
In the experiments, we generated a discrete rough terrain with randomized configurations, which included curbs and irregular cobblestone obstacles.
As shown in Table~\ref{tab:ablation_vectors}, our {DAMP} framework demonstrates strong environmental adaptability. While the HIM framework produces plausible motions under the same AMP supervision, it struggles to maintain robust task execution. {DAMP w/o} $\mathcal{L}_{\text{denoise}}$ exhibits inferior performance. Although it relies on temporal feature memorization, it fails to extract informative latent representations of the robot’s intrinsic states and other advantageous cues, resulting in degraded reactive capability. {DAMP w/o} $\mathcal{L}_{\text{AMP}}$ and $r_{\text{style}}$ can adapt to terrain variations to some extent. However, without paradigm guidance, the policy tends to generate undesirable behaviors such as left–right asymmetry, which are detrimental to locomotion. Moreover, training becomes more sensitive to reward tuning, with stronger coupling between task reward components and a longer hyperparameter adjustment cycle. 
In addition, we found that incorporating AMP significantly accelerated locomotion training. By leveraging reference motion data and an adversarial discriminator, the policy achieved faster convergence.

Beyond task-level performance comparisons, Fig.~\ref{predict_lin} illustrates the prediction error of the linear velocity estimator. Over 500 inference steps, larger deviations are observed during the transient startup phase due to rapid state changes. As locomotion stabilizes, both tracking lag and estimation error diminish, demonstrating the estimator’s ability to capture consistent latent dynamics during steady-state motion.
\begin{table}[t]
\centering
\caption{Average terrain level achieved by different frames in the simulation on different terrains.}
\label{tab:ablation_vectors}
\renewcommand{\arraystretch}{1.1}
\begin{tabular}{lcccc}
\toprule[1.25pt]
Method & Stairs & Slope & Discrete Rough \\
\midrule
\textbf{\textit{DAMP (ours)}} & \textbf{5.76} & \textbf{5.84} & \textbf{5.89} \\
\textit{DAMP w/o} $\mathcal{L}_{AMP}$ and $r_{\text{style}}$ & 4.74 & 4.89 & 4.32  \\
\textit{HIM}  & 4.54 & 4.65 & 4.33 \\
\textit{DAMP w/o} $\mathcal{L}_{\text{denoise}}$ & 3.38 & 3.36 & 3.78 \\

\bottomrule[1.25pt]
\end{tabular}
\end{table}

\subsection{ Real-World Experiments}
\begin{figure}[htbp]
    \centering
    \includegraphics[width=\linewidth]{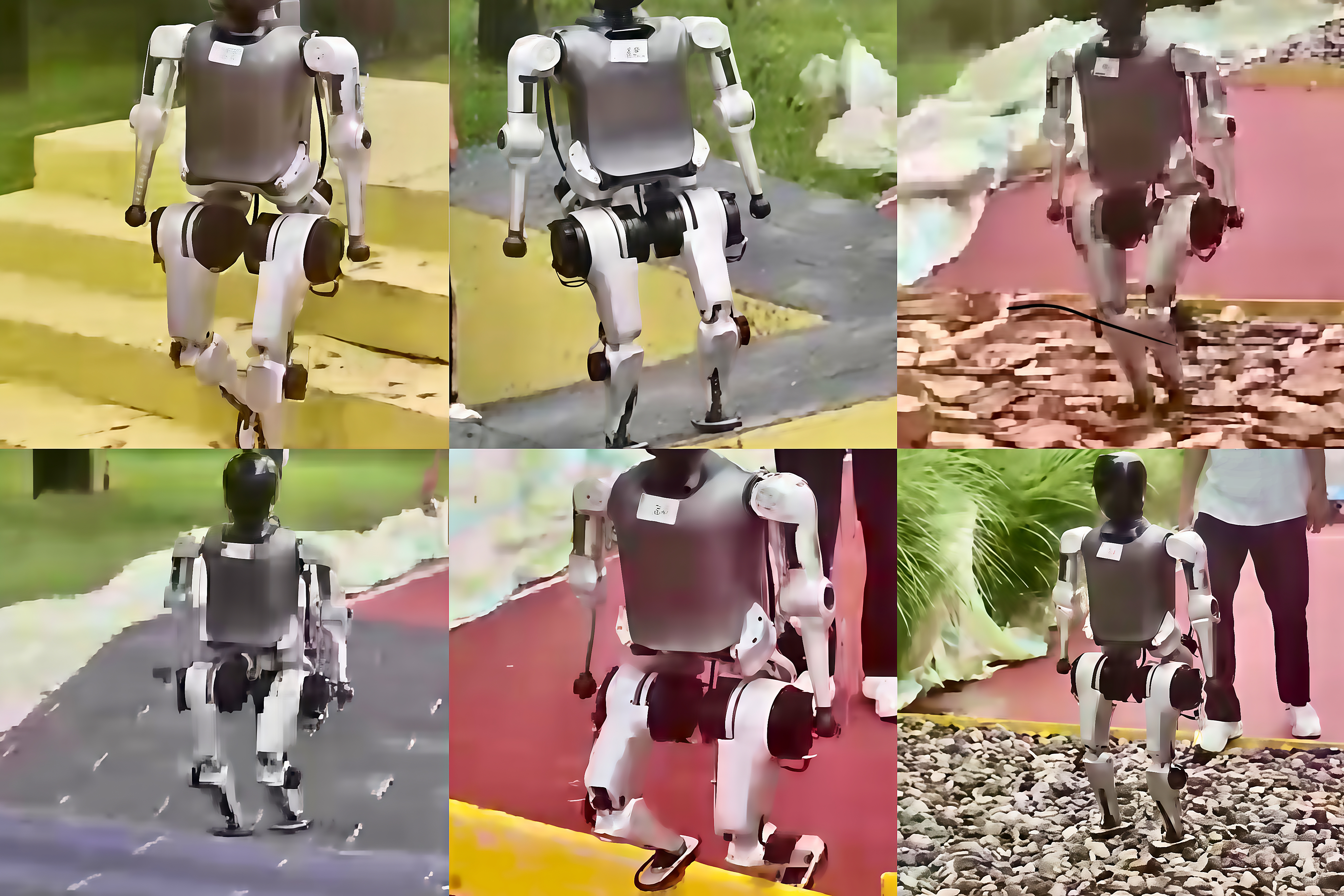}
    \caption{This strategy was tested across varied terrain, presented from left to right and top to bottom as follows: 
a $15\,\mathrm{cm}$ step, a $20^\circ$ staggered ramp, an irregular flagstone path, 
a $15^\circ$ continuous slope, a $20\,\mathrm{cm}$ kerb, and an irregular cobblestone obstacle.
}
    \label{fig-real}
\end{figure}
We compared the success rates achieved by DAMP on the designed terrains with those obtained from real-world ablation experiments to evaluate its stability and maneuverability across diverse terrains. Each method was evaluated over 20 trials for each terrain type and difficulty level, including stairs, ramps, and discrete rough terrains.

\begin{figure*}[!t]
    \centering
    \includegraphics[width=\textwidth]{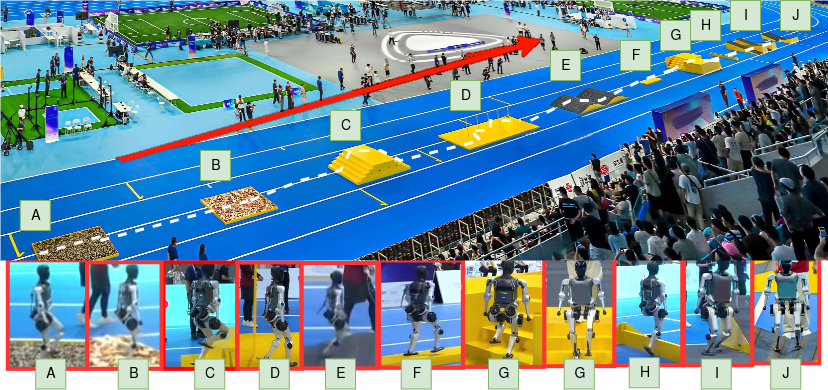}
    \caption{Overall terrain map of the competition arena. A–J denote terrain types. The N2 robot successfully traverses all obstacles in the real-world competition arena.}
    \label{fig-overall}
\end{figure*}
\begin{figure}[htbp]
    \centering
    \includegraphics[width=\linewidth]{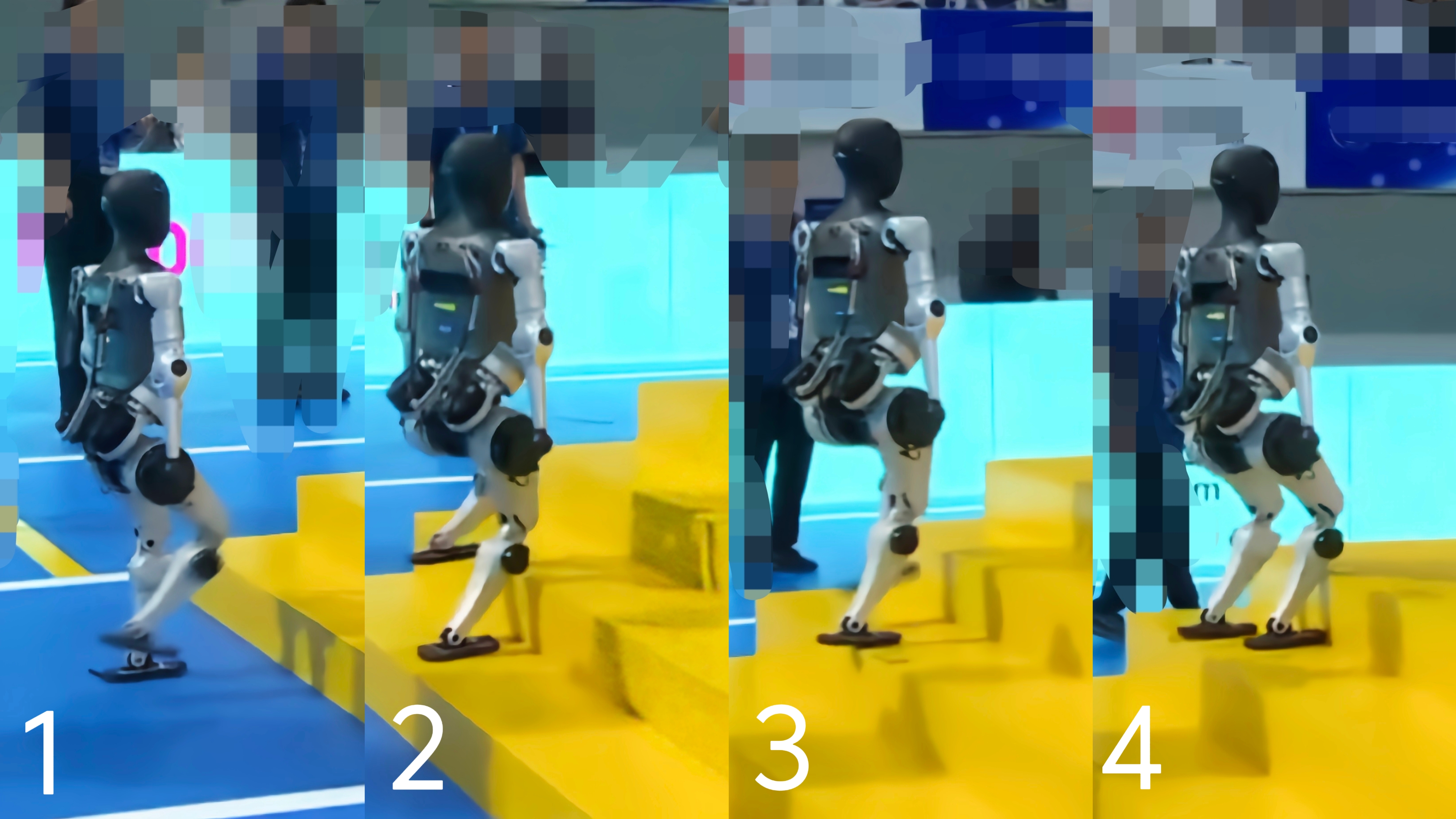}
    \caption{Real-world assessment of adaptive motion of a humanoid robot on stairs.
}
    \label{fig-real-stair}
\end{figure}

As shown in Fig.~\ref{success_rate}, we evaluate performance using the traversal success rate as the primary metric. In this ablation study, DAMP demonstrates superior capability across diverse terrains, benefiting from implicit state estimation and the stabilizing effect introduced by adversarial motion priors. While HIM exhibits difficulty in progressing to higher terrain difficulty levels in simulation, its traversal success rates in real-world experiments indicate a certain degree of robustness. However, it remains less competitive in task-specific evaluation metrics. Although {DAMP w/o} $\mathcal{L}_{AMP}$ and $r_{\text{style}}$ still achieves reasonable performance, it exhibits pronounced oscillations and undesirable behaviors—such as excessive leg lifting and ankle collisions—when traversing highly complex terrains. In contrast, {DAMP w/o} $\mathcal{L}_{\text{denoise}}$ performs significantly worse. A standalone LSTM module is insufficient to extract reliable latent representations from explicit proprioceptive observations alone, resulting in degraded task alignment, frequent instability, and failure to successfully traverse challenging terrains.

In \autoref{fig-real}, we illustrate representative traversal behaviors
across diverse real-world terrains.
The robot negotiates discrete elevation changes,
continuous slopes, and irregular footholds,
demonstrating stable contact transitions and recovery behaviors.
Furthermore, we deployed the learned policy to participate in the 100-meter obstacle course of the World Humanoid Robot Games 2025, completing the full challenge. The obstacle types follow the official specifications described in \cite{whrgoc2025}, providing a standardized benchmark for evaluating real-world terrain robustness. As shown in Fig.~\ref{fig-overall}, the N2 robot successfully traverses all terrain types (A–J) in the competition arena. Fig.~\ref{fig-real-stair} further demonstrates the robot’s adaptive locomotion capability when negotiating real-world stair obstacles.

\section{CONCLUSIONS}
We propose a lightweight and unified framework that estimates latent implicit states from explicit observations, driven by task objectives and enhanced with imitation learning. The framework further incorporates LCP-based constraints during policy optimization to improve robustness and stability. Simulation ablation studies demonstrate that both historical context for latent state inference and imitation guidance are essential for maintaining stability and task performance. Real-world experiments further validate the system’s agile and adaptive locomotion capabilities across stairs, slopes, and discrete terrains. Overall, these results indicate that integrating proprioceptive-to-privileged state inference with reinforcement learning provides a robust and effective pathway toward reliable humanoid locomotion in unstructured environments.





\section*{ACKNOWLEDGMENT}
This research was supported by the National Science and Technology Major Project of China (2025ZD1606201), the National Natural Science Foundation of China (62373223), and the Natural Science Foundation of Shandong Province (ZR2024ZD06).


\printbibliography

\end{document}